\documentclass[10pt,twocolumn,letterpaper]{article}

\usepackage{cvpr}              
\definecolor{cvprblue}{rgb}{0.21,0.49,0.74}
\usepackage[pagebackref,breaklinks,colorlinks,allcolors=cvprblue]{hyperref}
\usepackage{graphicx}
\usepackage{algpseudocode}
\usepackage{amsmath}
\usepackage{amssymb}
\usepackage{mathtools}
\usepackage{amsthm}
\usepackage{color, colortbl}
\usepackage{algorithm}
\usepackage{algpseudocode}
\usepackage{varwidth}
\definecolor{LightCyan}{rgb}{0.88,1,1}
\definecolor{LightRed}{rgb}{1,0.88,0.88}
\definecolor{Gray}{rgb}{0.8,0.8,0.8}
\definecolor{LightGray}{rgb}{0.92,0.92,0.92}
\definecolor{LightPurple}{RGB}{226, 225, 254}
\usepackage{multirow}
\usepackage{xcolor}
\usepackage{caption}
\usepackage{comment}
\usepackage{wrapfig}
\usepackage{titletoc}
\usepackage{diagbox}
\usepackage{graphicx}
\newtheorem{theorem}{Theorem}

\newtheorem*{theorem*}{Lemma}

\usepackage[margin=1in]{geometry}
\usepackage{amsmath,amssymb,amsthm,mathtools}
\usepackage{bm}
\usepackage{bbm}
\usepackage{microtype}
\usepackage{graphicx}
\usepackage{booktabs}
\usepackage{enumitem}
\usepackage{hyperref}
\usepackage{graphicx}
\usepackage{tikz}
\usetikzlibrary{positioning}

\theoremstyle{plain}

\def\paperID{46270} 
\def\confName{CVPR}
\def\confYear{2026}

\usepackage{amsmath}
\usepackage{tikz}
\usetikzlibrary{positioning}

\title{Uncertainty-Aware Knowledge Distillation for Multimodal Large Language Models}

\author{
Jingchen Sun$^{1,2}$\thanks{Work done during an internship at NEC Laboratories America.},
Shaobo Han$^{1}$\thanks{Project Lead},
Deep Patel$^{1}$,
Wataru Kohno$^{1}$,
Can Jin$^{3}$,
Changyou Chen$^{2}$\\[4pt]
$^{1}$NEC Laboratories America, Inc., USA \quad
$^{2}$University at Buffalo, SUNY \quad
$^{3}$Rutgers University
}

\begin{document}
{
 \maketitle
\begin{abstract}

Knowledge distillation establishes a learning paradigm that learns from both data supervision and teacher guidance. 
However, the optimal balance between learning from data and learning from the teacher is hard to determine, as some samples are data-noisy while others are teacher-uncertain. This raises a pressing need to adaptively balance data and teacher supervision.
We propose \textbf{Beta}-weighted \textbf{K}nowledge \textbf{D}istillation \textbf{(Beta-KD)}, an uncertainty-aware distillation framework that adaptively modulates how much the student rely on the teacher guidence. 
Specifically, we formulate teacher--student learning from a unified Bayesian perspective and interpret teacher supervision as a Gibbs prior over student activations. This yields a closed-form, uncertainty-aware weighting mechanism and supports arbitrary distillation objectives and combination. 
Extensive experiments are conducted on multimodal VQA benchmarks by distilling a student Vision-Language Model (MobiVLM and LLaVA) from a large teacher VLM. The results demonstrate that Beta-KD consistently outperforms existing knowledge distillation methods. Code is available at \url{https://github.com/Jingchensun/beta-kd}.
\end{abstract}    
\section{Introduction}

\begin{figure}[htb!]
    \centering
    \includegraphics[width=0.475\textwidth]{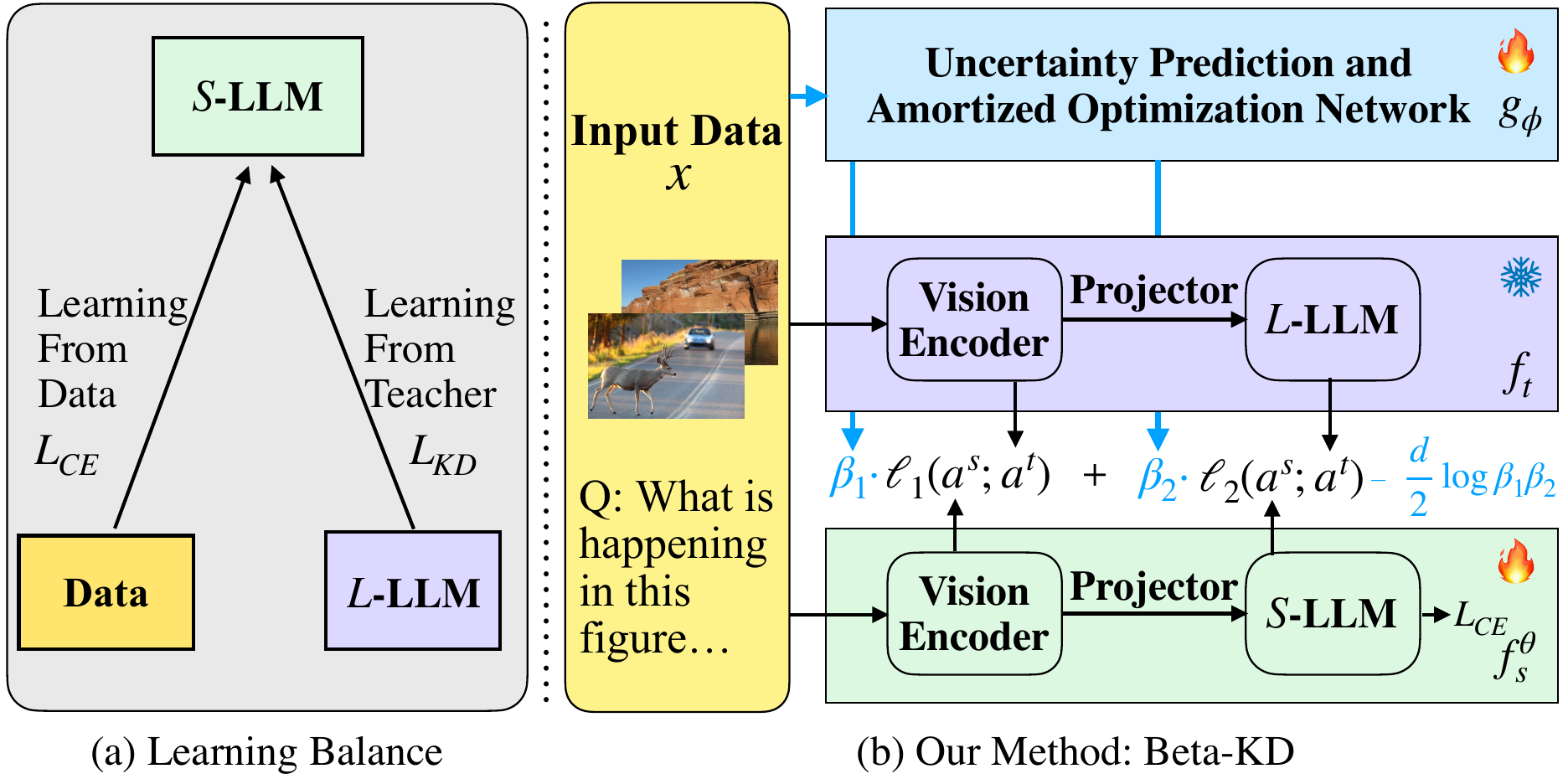}
    \caption{\textbf{Overview of the proposed Beta-KD framework.} 
    (a) Conventional KD is hard to balance the learning from data and the learning from teacher signals.  
    (b) Our method introduces an uncertainty-aware weighting framework by recognizing teacher supervision as a Gibbs prior, which naturally induces the prediction of the weights $\beta_1$ and $\beta_2$ through an amortized optimization network. The predicted uncertainty weights dynamically modulate the learning strength between teacher and student alignment, enabling adaptive balancing without manual hyperparameter tuning.
    \label{fig:motivation}
    }\vspace{-20pt}
\end{figure}

Recent advances in multimodal large language models (MLLMs), such as LLaVA~\citep{liu2024llava}, MiniGPT-4~\citep{zhu2023minigpt4}, and Qwen-VL~\citep{bai2023qwen}, have demonstrated impressive cross-modal understanding \cite{craft} and reasoning capabilities. However, as model scales continue to grow, efficiency and deployability have become major challenges, motivating research toward more compact and efficient MLLMs \cite{mobilevlm-v2, jincan2, prompt}. Knowledge Distillation (KD) has long served as an effective framework for transferring knowledge from a large, well-trained \emph{teacher} model to a compact \emph{student} model, enabling smaller models to achieve comparable performance with substantially reduced computational and memory costs~\cite{model-compress}. Early KD studies primarily focused on discriminative tasks such as image classification, where the teacher’s final-layer logits were used to guide the student’s predictions~\cite{hinton, model-compress, do-deep}. Later works extended KD to intermediate representations, including feature maps~\cite{fitnets, patient, tinybert}, attention maps~\cite{mobilebert, minilm}, and teacher assistants~\cite{ta}, aligning multiple layers between the teacher and student. 

In generative modeling, KD has also proven highly effective for building efficient language models \cite{distilbert, tinybert, minilm, mobilebert, jincan1}. MiniLLM~\cite{minillm} mitigates teacher–student distribution mismatch via a reverse-KL objective~\cite{distillm}, while GKD~\cite{onpolicy} performs knowledge transfer on student-generated samples, allowing the model to learn from its own inference trajectories and teacher feedback. DistiLLM~\cite{distillm}, TAID~\cite{taid}, and DDK~\cite{ddk} further improve efficiency through mechanisms that consider training dynamics, temporal progression, and domain disparities between teacher and student. For multimodal and vision-language models, Align-KD~\cite{alignkd} and LLaVA-KD~\cite{llavakd} extend this paradigm to distill cross-modal alignment knowledge, preserving teacher–student correspondence in visual-textual representation spaces. 

However, distilling multimodal large language models remains challenging. 
The first issue lies in how to balance learning from data and learning from the teacher model, as illustrated in Figure~\ref{fig:motivation} (a). The cross-entropy loss corresponds to learning from data, while the KL divergence measure learning from the teacher’s predictive distribution. In MLLMs, there are usually additional channels such as visual or textual feature matching, where the feature-level alignment loss enforces learning from teacher's latent representations. Balancing these heterogeneous supervisory signals is inherently non-trivial, since each exhibits different scales, gradients, and optimization dynamics. This challenge is further amplified by the large capacity gap between the teacher and student, which causes discrepancies in the scale and variance of their logits and hidden representations, leading to imbalanced learning objectives.

To address this issue, we propose \textbf{Beta-weighted Knowledge Distillation (Beta-KD)}, an uncertainty-aware knowledge distillation framework that to adjust the learning signal from teacher or from data in an adaptive way. We model the student's activations as the data likelihood and the teacher's supervision as a Gibbs prior, framing the distillation process as amortized Maximum a Posteriori (MAP) estimation. 
This yields a Gibbs posterior whose mode corresponds to minimizing a standard distillation loss augmented with an uncertainty-dependent precision term. 
By applying the Laplace approximation, we derive a closed-form weighting mechanism that introduces both task-level and instance-level uncertainty, enabling adaptive, data-driven loss balancing and eliminating the need for manual weighting search. The uncertainty parameters are optimized through a neural network parameterization.

We distill a 1.7B-parameter student from MobileVLM-7B and evaluate Beta-KD under both two-loss and three-loss settings.
Across all configurations, Beta-KD consistently improves distillation performance.
Task-level uncertainty weighting achieves a substantial gain on {ScienceQA}, improving VQA accuracy by up to {+4.0} absolute points, while instance-level uncertainty yields an even larger {+4.7}-point improvement.
Training-dynamics visualizations further show faster convergence, smoother optimization, and closer teacher--student logit alignment.
When scaled to a larger transfer set and evaluated on six multimodal benchmarks, our best configuration delivers \emph{consistent} improvements with up to a {+2.0}-point \emph{average} gain, establishing a new state-of-the-art in multimodal knowledge distillation.

Our main contributions are as follows:
\begin{itemize}
    \item \textbf{We introduce a Bayesian inference perspective on knowledge distillation} based on \emph{teacher-informed Gibbs priors} on student activations. 
    This formulation unifies existing KD methods under a probabilistic framework that naturally incorporates uncertainty modeling.  
    We show that KD training can be viewed as finding the MAP solution for student activations via amortized neural inference.

    \item \textbf{We derive an uncertainty-aware weighting mechanism} using the \emph{Laplace approximation}.  
    This closed-form solution enables adaptive instance-level and task-level loss balancing through an uncertainty network.  
    In multimodal LLMs, it selectively preserves informative teacher signals while reducing noise and improving data quality during distillation.

    \item \textbf{We study the various design choices of incorporating teacher prior knowledge}, including ablations on both the \emph{activation locations} (e.g., logits vs.\ probabilities) and the \emph{activation formulations} (e.g., different KL-based losses). We find that under various loss combinations and experimental settings, both task-level and instance-level uncertainty weighting consistently improve model performance on six large-scale VQA benchmarks. 
\end{itemize}

\section{Related Work}

\paragraph{KD in Multimodal LLMs.}

Recent studies have explored diverse KD objectives, including Reverse KL (RKL)~\cite{minillm}, Skew KL/RKL~\cite{distillm}, and $\alpha$–$\beta$ divergence-based losses~\cite{abkd}, along with engineered variants for LLMs~\cite{distillm, taid}. 
While KD has been widely studied in unimodal LLMs (e.g., MiniLLM~\cite{minillm}, DistiLLM~\cite{distillm}, TAID~\cite{taid}), multimodal distillation remains challenging due to the multi-channel loss balancing issue (e.g, vision tokens, textual features, and cross-modal embeddings). Our work aims to establish a unified Bayesian inference framework that adaptively adjusts the weights of multiple loss channels in multimodal LLM distillation. Compare to related work BayesKD \cite{bayeskd}, which provides a \emph{statistical interpretation} of why KD works and estimates the uncertainty of the model parameters $\theta$. Our work Beta-KD formulates knowledge transfer in distillation as an amortized Bayesian inference problem over student activations, where uncertainty is used to weight the contributions of different loss terms.

\vspace{-10pt}
\paragraph{Uncertainty-based Loss Balancing.}
The loss balancing problem in KD  closely parallels that in multi-task learning (MTL)~\citep{multitask, uncertainties}.
Gradient-based methods such as GradNorm~\citep{chen2018gradnorm} and PCGrad~\citep{yu2020gradient} adapt task weights via gradient normalization, but they often underperform in practice and are impractical for multimodal LLMs due to their computational overhead.
The most closely related work is the multi-task weighting method proposed by Kendall \& Gal~\citep{multitask}, originally developed for image classification and regression tasks.

However, Kendall \& Gal~\citep{multitask} assume that task losses arise from Gaussian likelihoods and derive task-level weights via maximum likelihood estimation with asymptotic approximations.
In contrast, we generalize uncertainty-based weighting to \emph{arbitrary distillation losses} by modeling the teacher--student discrepancy as a Gibbs prior, covering FKL~\cite{hinton}, RKL~\cite{rkl}, SFKL~\cite{skew-fkl-rkl}, etc., and enables instance-level adaptive loss balancing across multiple KD objectives.



\section{Uncertainty-Aware Knowledge Distillation}

\subsection{Preliminaries}
\label{sec:preliminaries}
We consider the standard \textbf{knowledge distillation (KD)} framework between a fixed \emph{teacher model} $f_t$ and a parameterized \emph{student model} $f_s$ with trainable parameters $\theta$.  
Given an input sequence $x$ and its target output $y=(y_1,\dots,y_{L_y})$, both teacher and student produce token-level logits over the shared vocabulary $\mathcal{V}$:
$\mathbf{z}_t(x,y_{<n}),\;
\mathbf{z}_s(x,y_{<n};\theta) \in \mathbb{R}^{|\mathcal{V}|},$
where $y_{<n}=(y_1,\dots,y_{n-1})$.  
After applying temperature-scaled softmax, the probability distributions are:
$\mathbf{p}_t^{\tau_t} = \operatorname{softmax}\!\big(\mathbf{z}_t / \tau_t\big), 
\mathbf{p}_s^{\tau_s} = \operatorname{softmax}\!\big(\mathbf{z}_s / \tau_s\big).$



\textbf{Cross Entropy (CE).}
In autoregressive LM, the student maximizes the sequence likelihood, equivalently minimizing cross-entropy against the hard label:
\[
\mathcal{L}_{\mathrm{CE}}
= -\frac{1}{L_y}\sum_{n=1}^{L_y}\sum_{k=1}^{|\mathcal{V}|} e_k(y_n)\,\log p^{\tau_s}_{s,k}(y_n\mid x,y_{<n};\theta),
\]
where $\mathbf{e}$ is the one-hot label of the ground-truth token $y_n$.

\textbf{Knowledge Distillation (KD).}
KD replaces the hard label with the teacher's soft target and trains the student to match the teacher distribution:
\[
\mathcal{L}_{\mathrm{KD}}
= \frac{1}{L_y}\sum_{n=1}^{L_y}
\mathbb{D}\!\left(
\mathbf{p}_t^{\tau_t}
\,\big\|\,
\mathbf{p}_s^{\tau_s}(\cdot \mid x,y_{<n};\theta)
\right),
\]
where $\mathbb{D}(\cdot\|\cdot)$ denotes a divergence such as FKL~\cite{hinton}, RKL~\cite{rkl}, SFKL \cite{skew-fkl-rkl}.

\textbf{Final Training Objective.}
The overall training objective combines the CE loss and the distribution-level KD loss:
\begin{equation}
\label{eq:total-loss}
\mathcal{L}_{\text{total}}
=
\mathcal{L}_{\mathrm{CE}}
+ 
\lambda\,\mathcal{L}_{\mathrm{KD}},
\end{equation}
where $\lambda$ controls the relative contribution of the distillation term.

\textbf{How to choose the optimal $\lambda$?} 
Determining the optimal value of $\lambda$ in Eq.~\eqref{eq:total-loss} is challenging, especially for multimodal LLM distillation, where the KD objective often comprises multiple sub-losses, e.g, 
$
\mathcal{L}_{\text{total}}
=
\mathcal{L}_{\mathrm{CE}}
+ 
\lambda_{1}\,\mathcal{L}_{\mathrm{KD_1}}
+
\lambda_{2}\,\mathcal{L}_{\mathrm{KD_2}}
$, then to find the best combination of $\lambda_{1}$ and $\lambda_{2}$ is not trivial.  These loss terms exhibit different scales, gradient magnitudes, and optimization dynamics, making it particularly difficult to balance their supervision signals. While grid search is commonly used, it is impractical for large-scale LLMs due to the expensive computing cost.
To address this, we interpret this weighting hyperparameter as reflecting the \emph{reliability of teacher supervision}, and propose an empirical Bayes-based approach that automatically adjusts the relative weight, enabling efficient and adaptive loss balancing. 
Following common ML/CV notation, we use 
$\beta$ to denote the uncertainty weighting parameter corresponding to the relative weight 
$\lambda$.

\subsection{A Bayesian View of Knowledge Distillation}
\label{sec:bayesian-gibbs}

We interpret the language modeling data-generating process in \textbf{Figure~\ref{fig:data_chain}} from a Bayesian inference viewpoint.
Given an input $x$, the student network produces intermediate activations, which may refer to feature representations $\mathbf{f}^s$, logits $\mathbf{z}^s$, or output probabilities $\mathbf{q}^s$.
For notational convenience, we denote the chosen student activation at a given level as $a^s \in \{\mathbf{f}^s,\mathbf{z}^s,\mathbf{q}^s\}$, and the corresponding teacher activation as $a^t \in \{\mathbf{f}^t,\mathbf{z}^t,\mathbf{p}_t^\tau\}$.
The student activation is induced by the model parameters $\theta$ (i.e., $a^s=a^s(x;\theta)$), and the teacher information $(a^t,\beta)$ acts as an \emph{informed prior} that guides the student activation.
We estimate $a^s$--equivalently, the parameters $\theta$ that induce it--via Maximum A Posteriori (MAP) inference.

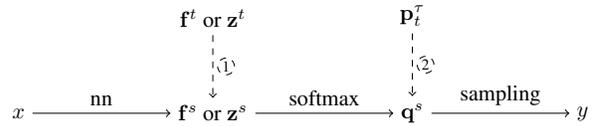
\begin{figure}[htbp]
\centering
\resizebox{\columnwidth}{!}{%
\begin{tikzpicture}[node distance=2.2cm, baseline=(current bounding box.center)]
    \node (x) {$x$};
    \node[right=of x] (fzs) {$\mathbf{f}^{s}$ or $\mathbf{z}^{s}$};
    \node[right=of fzs] (pi) {$\mathbf{q}^{s}$};
    \node[right=of pi] (y) {$y$};

    \draw[->] (x) -- node[above]{nn} (fzs);
    \draw[->] (fzs) -- node[above]{softmax} (pi);
    \draw[->] (pi) -- node[above]{sampling} (y);

    \node[above=1.0cm of fzs] (ftzt) {$\mathbf{f}^{t}$ or $\mathbf{z}^{t}$};
    \draw[dashed,->] (ftzt) -- (fzs)
        node[right, midway, xshift=2pt, draw, circle, inner sep=0.6pt, font=\scriptsize]{1};

    \node[above=1.0cm of pi] (pit) {$\mathbf{p}_t^\tau$};
    \draw[dashed,->] (pit) -- (pi)
        node[right, midway, xshift=2pt, draw, circle, inner sep=0.6pt, font=\scriptsize]{2};
\end{tikzpicture}%
}
\caption{\textbf{Language modeling chain with teacher guidance.}
Given input $x$, the student network produces activations $\mathbf{f}^s$ or $\mathbf{z}^s$, which are mapped to probabilities $\mathbf{q}^s$ via softmax and then sampled to generate output $y$.
Dashed arrows indicate teacher supervision injected at (1) the feature/logit level ($\mathbf{f}^t$ or $\mathbf{z}^t$) or (2) the probability level ($\mathbf{p}_t^\tau$).}
\label{fig:data_chain}
    \vspace{-20pt}
\end{figure}

\paragraph{Teacher-Informed Gibbs Prior.}
Motivated by statistical physics, we cast knowledge distillation as an \emph{energy-based} learning problem.
In our setting, we treat the \emph{teacher--student discrepancy} $\ell(a^s;a^t)$ as the energy measuring how well the student activation $a^s$ aligns with the teacher signal $a^t$.
This choice is natural because it converts \emph{matching the teacher} into an energy-minimization principle, and the resulting Gibbs form~\cite{gibbs} provides a principled probabilistic prior that (i) assigns \emph{lower} probability to large teacher--student discrepancies and \emph{higher} probability to small discrepancies, and (ii) exposes an explicit \emph{concentration} parameter $\beta$ that controls the strength of supervision.
We define the (unnormalized) teacher-informed prior as
\begin{align}
\tilde p(a^s \mid a^t, \beta)
\;\propto\;
\exp\!\big[-\beta\,\ell(a^s; a^t)\big],
\qquad \beta>0,
\label{eq:gibbs-prior-unnorm}
\end{align}
Intuitively, a larger $\beta$ (lower temperature) corresponds to a sharper, more confident prior that aligns the student more tightly to the teacher, whereas a smaller $\beta$ yields a smoother, more uncertain prior. To obtain a proper probability distribution, we normalize by the partition function
\begin{align}
Z_\beta(a^t)
\;=\;
\int \exp\!\big[-\beta\,\ell(a^s; a^t)\big]\,da^s,
\label{eq:intergal-z}
\end{align}
yielding the teacher-induced prior
\begin{align}
p(a^s \mid a^t, \beta)
=
\frac{1}{Z_\beta(a^t)} \exp\!\big[-\beta\, \ell(a^s; a^t)\big].
\label{eq:gibbs-prior}
\end{align}
In high-dimensional spaces, $Z_\beta(a^t)$ is typically intractable, which motivates the approximation in the sequel.
Here, $\ell(a^s; a^t)$ can take different forms of energy functions, such as Forward KL (FKL)~\cite{hinton} or Reverse KL (RKL)~\cite{rkl}, which correspond to distinct assumptions about how the student aligns with the teacher.  
We further conduct a comparative study exploring alternative energy formulations beyond KL-based losses, such as Mean Squared Error (MSE) \cite{abkd} and Cosine Distance \cite{cosine}, to model the teacher–student relation more effectively in probability space. 
A qualitative comparison of these energy functions is illustrated in \textbf{Figure~\ref{fig:energy_losses}}.
\label{sec:method}

\begin{figure}[htb!]
    \centering
    \includegraphics[width=0.45\textwidth]{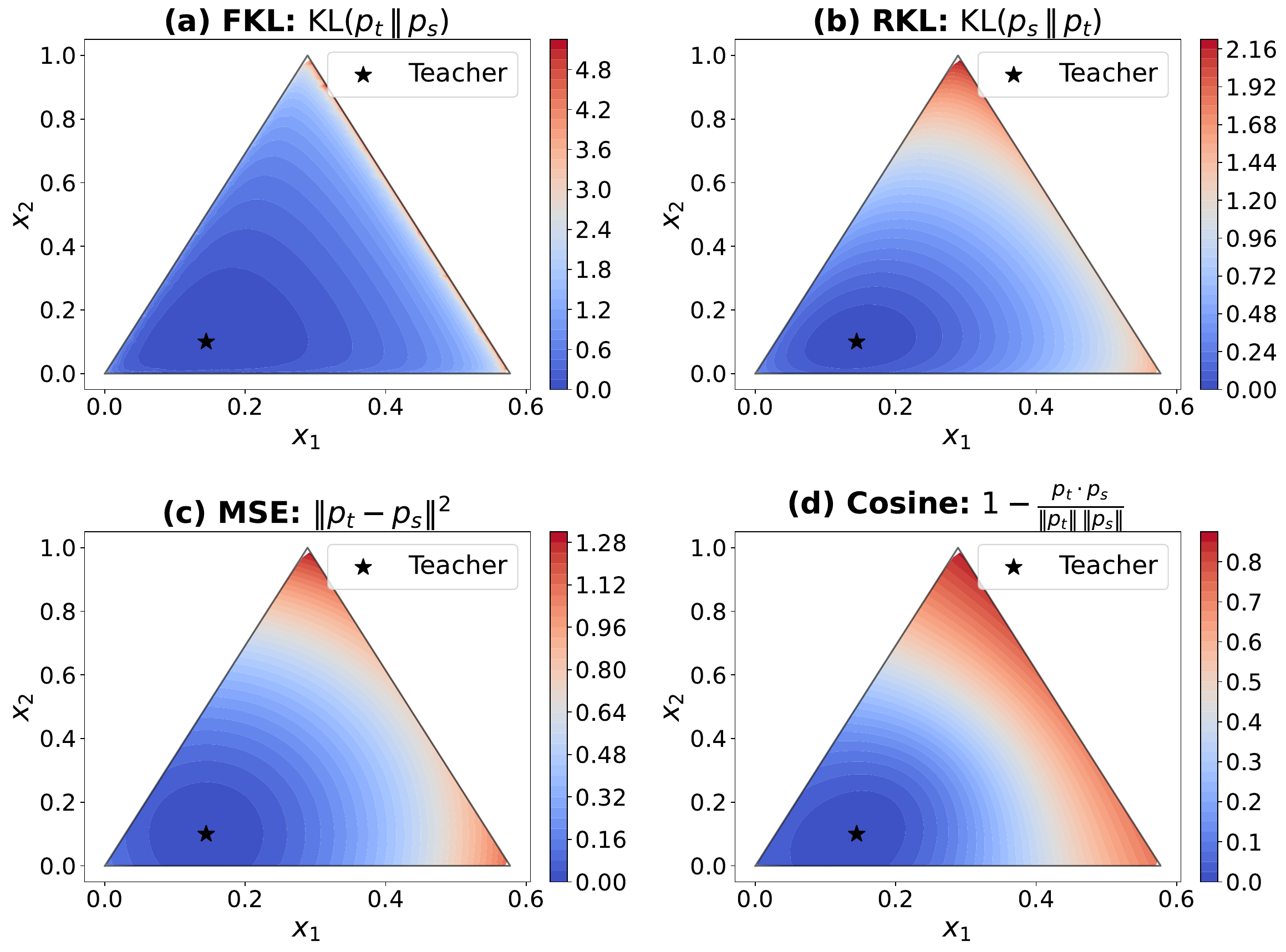}
    \caption{
Visualization of four representative knowledge distillation losses in the probability simplex.}
    \label{fig:energy_losses}
    \vspace{-20pt}
\end{figure}

\paragraph{MAP of Student Activation.}

We formulate knowledge distillation (KD) from a Bayesian perspective by modeling the student activation $a^s$ as a latent variable guided by teacher information $(a^t,\beta)$.  
The central modeling assumption in KD is that the teacher does not directly determine the output $y$, but instead influences it only through the student representation $a^s$. Formally, we assume
\begin{equation}
p(y\mid a^s,a^t,\beta)=p(y\mid a^s),
\label{eq:cond-indep}
\end{equation}
which is equivalent to the conditional independence relation
\[
y \perp (a^t,\beta)\mid a^s.
\]

Under this assumption, the joint distribution over the student activation and the output can be factorized using the probability chain rule as
\begin{equation}
p(y,a^s\mid a^t,\beta)
=
p(y\mid a^s)\,p(a^s\mid a^t,\beta).
\label{eq:joint-factor}
\end{equation}
This corresponds to the data generative structure $(a^t,\beta)\rightarrow a^s \rightarrow y$,
where the teacher information shapes the student activation, which subsequently generates the output.

Applying Bayes' theorem, the posterior distribution over the student activation is given by
\begin{equation}
p(a^s\mid y,a^t,\beta)
=
\frac{p(y\mid a^s)\,p(a^s\mid a^t,\beta)}
{p(y\mid a^t,\beta)},
\label{eq:posterior}
\end{equation}
where the marginal likelihood (evidence)
\[
p(y\mid a^t,\beta)
=
\int p(y\mid a^s)\,p(a^s\mid a^t,\beta)\,da^s
\]
serves as a normalization constant independent of $a^s$.

We then perform Maximum A Posteriori (MAP) inference:
\begin{align}
a^{s\ast}
&=\arg\max_{a^s}\; p(y\mid a^s)\,p(a^s\mid a^t,\beta), \nonumber\\
&=\arg\min_{a^s}\; \Big[-\log p(y\mid a^s)-\log p(a^s\mid a^t,\beta)\Big].
\label{eq:map}
\end{align}

The first term is the data likehood and the second term is the teacher informed prior. Substituting the Gibbs prior in Eq.~\eqref{eq:gibbs-prior} into Eq.~\eqref{eq:map}, we obtain
\begin{equation}
-\log p(a^s\mid a^t,\beta)
=
\beta\,\ell(a^s;a^t)
+
\log Z_\beta(a^t),
\label{eq:gibbs-neglog}
\end{equation}
and hence the MAP objective is
\begin{equation}
\min_{a^s}
\underbrace{-\log p(y\mid a^s)}_{\text{Cross-Entropy (CE)}}
+
\beta\,\underbrace{\ell(a^s;a^t)}_{\text{Distillation Loss (KD)}}
+
\log Z_\beta(a^t).
\label{eq:map-gibbs}
\end{equation}
This shows that MAP inference over the student activation is equivalent to minimizing the teacher--student distillation discrepancy, as we stated in Theorem \ref{thm:energy-bayes}.
\begin{theorem}[Energy--Bayes Equivalence]
\label{thm:energy-bayes}
Let the teacher-informed prior be a Gibbs distribution
\[
p(a^s \mid a^t,\beta)
=
\frac{1}{Z_\beta(a^t)}\exp\!\big(-\beta\,\ell(a^s;a^t)\big),
\qquad \beta>0,
\]
and assume $p(y\mid a^s,a^t,\beta)=p(y\mid a^s)$.
Then maximizing the posterior $p(a^s\mid y,a^t,\beta)$ is equivalent to minimizing the knowledge distillation objective
\[
\mathcal{J}(a^s)
=
-\log p(y\mid a^s)
+
\beta\,\ell(a^s;a^t)
+
\log Z_\beta(a^t).
\]
In particular, since $a^s$ is deterministically induced by $\theta$ via $a^s=a^s(x;\theta)$, optimizing the student activation corresponds to minimizing this objective w.r.t. $\theta$.
\end{theorem}

\paragraph{Laplace Approximation.}
The partition function in Eq.~\eqref{eq:gibbs-prior} is generally intractable. 
To obtain a tractable dependence on $\beta$, we apply \emph{Laplace's method} and approximate the integral by a local Gaussian expansion around the energy minimizer.
Let $a^\star=\arg\min_{a^s}\ell(a^s;a^t)$ be a (local) minimizer and define the Hessian at $a^\star$:
\begin{equation}
H=\nabla^2_{a^s}\ell(a^s;a^t)\big|_{a^s=a^\star}\succ0,\qquad d=\dim(a^s).
\nonumber
\end{equation}
Using a second-order Taylor approximation around $a^\star$,
\begin{equation}
\ell(a^s;a^t)\approx \ell(a^\star;a^t)+\tfrac{1}{2}(a^s-a^\star)^\top H (a^s-a^\star).
\label{eq:taylor_loss}
\end{equation}
Substituting Eq.~\eqref{eq:taylor_loss} into Eq.~\eqref{eq:intergal-z} turns the integrand into a Gaussian in $a^s$, yielding
\begin{equation}
Z_\beta(a^t)\approx \tilde{Z}_\beta(a^t)=
e^{-\beta \ell(a^\star;a^t)}(2\pi)^{\frac{d}{2}}\beta^{-\frac{d}{2}}|H|^{-\frac{1}{2}}.\nonumber
\end{equation}
Taking logarithms gives
\begin{equation}
\log \tilde{Z}_\beta(a^t)=
-\beta \ell(a^\star;a^t)-\frac{d}{2}\log\beta+\text{const},
\label{eq:laplace_logZ}
\end{equation}
where ``const'' collects terms independent of $\beta$.
For common  alignment energies used in KD, the minimum discrepancy satisfies $\ell(a^\star;a^t)=0$, so substituting Eq.~\eqref{eq:laplace_logZ} into the MAP objective in Eq.~\eqref{eq:map-gibbs} yields the tractable surrogate:
\begin{equation}
\min_{a^s}\;
-\log p(y\mid a^s)
+\beta\,\ell(a^s;a^t)
-\frac{d}{2}\log\beta .
\label{eq:surrogate_obj}
\end{equation}
The additional term $-\tfrac{d}{2}\log\beta$ is induced by normalizing the Gibbs prior; it prevents $\beta$ from trivially diverging and thus provides an explicit regularization for learning uncertainty-aware distillation strength. 

\paragraph{Amortized Optimization on $\beta$.}
Performing per-instance posterior optimization of $\beta$ (e.g., via inner-loop iterative updates) is computationally expensive. 
Instead, we adopt \emph{amortized optimization} and joiontly optimize the weighting factor $\beta$ with model parameters $\theta$. 
This can be viewed as learning a \emph{neural approximation to the posterior precision}, analogous to amortized inference in variational autoencoders \cite{inference}, where iterative inference is replaced by a fast learned mapping.

Specifically, we consider two uncertainty granularities:
\textbf{(i) task-level (homoscedastic) uncertainty}, where $\beta$ reduces to a set of \emph{directly learnable positive scalars} (e.g., $\{\beta_k\}_{k=1}^K$) shared across all samples for each task/channel $k$ and independent of the input $x$; and
\textbf{(ii) instance-level (heteroscedastic) uncertainty}, where $\beta$ is predicted by a lightweight network from the input:
\begin{equation}
\beta(x)=g_\phi\!\left(h(x)\right)>0,
\label{eq:beta_net}
\end{equation}
where $h(x)$ is a small feature extractor and $g_\phi(\cdot)$ enforces positivity (e.g., via $\mathrm{softplus}$), as illustrated in Figure~\ref{fig:motivation}.
We jointly optimize the student parameters $\theta$ and the uncertainty parameters $\phi$ via backpropagation:
\begin{equation}
\min_{\theta,\phi}\;
\mathcal{L}_{\mathrm{CE}}(\theta)
+g_{\phi}(h(x))\,\ell(\theta)
-\frac{d}{2}\log g_{\phi}(h(x)),
\label{eq:amortized_beta}
\end{equation}
which removes manual loss weighting and enables efficient batch training with data-dependent supervision strength.



\section{Experiments}

In the experimental section, we focus on addressing the following four research questions:

\begin{enumerate}
    \item \textbf{RQ1: The Design Choices of Energy-Bayes KD.}  
    What is the most effective \emph{energy representation}—in both formulation {(e.g., distill loss)} and location {(e.g., activation layer)}—for transferring knowledge between the teacher and the student models?

    \item { \textbf{RQ2: Effectiveness of Uncertainty Weighting.} How effective is the proposed \emph{uncertainty-aware weighting} framework when applied to a two-loss or multiple-loss setting (e.g., CE + KD or CE + KD +FD)?}
    \item \textbf{RQ3: How Uncertainty Weighting Works.}  
    How does the proposed \emph{uncertainty estimation} framework {operate} during multimodal knowledge distillation?

    \item \textbf{RQ4: Generalization Ability.}  
    Can the proposed method generalize effectively across different {multimodal datasets }and mainstream multimodal LLM architectures?
\end{enumerate}

\subsection{Experiment Setup}  

We use MobileVLM V2 7B \cite{mobilevlm-v2} as the teacher model and MobileVLM V2 1.7B \cite{mobilevlm-v2} as the student model, following prior works for fair comparison~\cite{alignkd, llavadi}. For Sections 4.2 to 4.4, we select a representative VQA task - ScienceQA \cite{lu2022sqa} and use the training set as the transfer set for distillation, with its test set used for evaluation. For Section 4.5, we expand the transfer set to 2.2M image–text pairs, including data from COCO~\cite{chen2015coco}, SBU~\cite{ordonez2011im2text}, Visual Dialog~\cite{das2017visualdialog}, ShareGPT4V~\cite{chen2023sharegpt4v}, SQA~\cite{lu2022sqa}, IConQA~\cite{lu2021iconqa}, TextVQA~\cite{singh2019textvqa}, VSR~\cite{liu2023vsr}, and VIGC~\cite{wang2024vigc}. The distilled models are then evaluated on six benchmark datasets: GQA~\cite{hudson2019gqa}, SQA~\cite{lu2022sqa}, TextVQA~\cite{singh2019textvqa}, MME~\cite{fu2023mme}, MMBench~\cite{liu2024mmbench}, and POPE~\cite{li2023pope}. During distillation, both the vision encoder and tokenizer are frozen, and only the language backbone is fine-tuned. 

\subsection{The Design Space of Energy-Bayes KD}

\begin{table}[htbp]
\centering
\caption{
\textbf{Comparison of different energy-based models for student–teacher knowledge transfer. }
All losses except CE are distillation losses. 
\emph{MSE-Logits} and \emph{Cosine-Logits} denote losses applied at the pre-softmax logit level, 
while \emph{MSE-Probs} and \emph{Cosine-Probs} are applied at the post-softmax probability level. 
Results on the ScienceQA dataset (averaged over three runs) show that Cosine-Probs achieves the best performance.
}
\label{tab:scienceqa_kl_variants}
\setlength{\tabcolsep}{1pt}
\fontsize{7}{9}\selectfont
\begin{tabular}{l|c|l|c}
\noalign{\hrule height 1pt} 
\rowcolor{Gray}
\textbf{Method} & \textbf{Acc Mean $\pm$ Std} & \textbf{Method} & \textbf{Acc Mean $\pm$ Std} \\ \hline
Cross Entropy (CE)   & 48.4 $\pm$ 1.4  & Adaptive KL \cite{adaptivekl} & 44.1 $\pm$ 2.4  \\
FKL  \cite{hinton}  & 45.7 $\pm$ 1.6  & CTKD \cite{ctkd}        & 37.8 $\pm$ 2.8  \\
RKL \cite{minillm} & 42.8 $\pm$ 3.2  & TAID \cite{taid}        & 47.0 $\pm$ 0.9  \\
Skew FKL \cite{distillm} & 45.7 $\pm$ 3.4  & MSE-Logits          & 4.9  $\pm$ 10.5 \\
Skew RKL \cite{distillm} & 45.7 $\pm$ 2.7  & MSE-Probs    & 28.2 $\pm$ 10.6  \\
JS \cite{js-tvd} & 45.8 $\pm$ 3.6  & Cosine-Logits       & 4.0  $\pm$ 1.0  \\
TVD \cite{js-tvd} & 38.7 $\pm$ 2.6  & \textbf{Cosine-Probs} & \textbf{47.2 $\pm$ 0.9 } \\
\noalign{\hrule height 1pt} 
\end{tabular}
\end{table}

\textbf{Results.}  
As mentioned in Section~\ref{sec:method}, we explore multiple formulations of the energy function to identify the most effective representation for modeling student–teacher knowledge transfer. The experimental results are summarized in {Table~\ref{tab:scienceqa_kl_variants}, leading to two key findings. \textbf{First, pre-softmax logit matching methods (\emph{MSE-Logits} and \emph{Cosine-Logits}) do not perform well {on the generative multimodal tasks evaluated in our experiments}. }
While prior works  have shown that logit-level matching \cite{compare-mse} can be equivalent or even superior to probability-level KL divergence in discriminative settings, we find this assumption does not hold for generative MLLMs. 
This observation is consistent with prior findings reported in~\cite{llavadi}. \textbf{Second, \emph{Cosine-Probs} achieves the best overall performance, outperforming various KL variants}, even the latest LLM distillation method, TAID \cite{taid}. 
We attribute this improvement to the scale-invariant nature of cosine distance, which emphasizes directional alignment and relative token ranking consistency rather than absolute probability scale alignment, similar findings also show in \cite{cosine}. 


\begin{table*}[htbp]
\centering
\caption{
Experimental results of \textbf{two-loss balancing} on the \textbf{ScienceQA} dataset. 
Each baseline combines Cross-Entropy (CE) with a KL-based distillation loss. 
\textbf{Manual} uses fixed weights between CE and KL based on their initial scales. 
\textbf{Beta-KD (Task)} models task-level uncertainty shared across all samples, 
while \textbf{Beta-KD (Instance)} models instance-level uncertainty adaptive to each input. VQA-Acc denotes the overall question–answering accuracy across all questions, whereas IMG-Acc measures the accuracy on the subset of questions whose explicitly include image inputs. Both strategies consistently enhance knowledge distillation performance across different loss functions.
}
\label{tab:scienceqa_uncertainty}
\setlength{\tabcolsep}{5.5pt}
\fontsize{7}{9}\selectfont
\begin{tabular}{lll|lll|lll}
\noalign{\hrule height 1pt} 
\rowcolor{Gray}
\textbf{Method }            & \textbf{VQA-Acc}     & \textbf{IMG-Acc}     & \textbf{Method}             & \textbf{VQA-Acc}     & \textbf{IMG-Acc}      & \textbf{Method}             & \textbf{VQA-Acc}     & \textbf{IMG-Acc}      \\ \hline
CE + FKL           & 48.2        & 54.7        & CE + JS            & 48.5        & 54.8         & CE + CTKD          & 48.6        & 55.0         \\
w/ Manual          & 48.6        & 54.9        & w/ Manual          & 49.4        & 56.3         & w/ Manual          & 49.1        & 55.1         \\
\rowcolor{LightCyan}
w/ Beta-KD (Task)     & 49.3 (\textcolor{red}{+0.7}) & 55.3 (\textcolor{red}{+0.4}) & w/ Beta-KD (Task)     & 50.5 (\textcolor{red}{+1.1}) & 58.1 (\textcolor{red}{+1.7})  & w/ Beta-KD (Task)     & 49.9 (\textcolor{red}{+0.8}) & 55.3 (\textcolor{red}{+0.3})  \\
\rowcolor{LightCyan}
w/ Beta-KD (Instance) & 51.8 (\textcolor{red}{+3.2}) & 61.1 (\textcolor{red}{+6.2}) & w/ Beta-KD (Instance) & 53.3 (\textcolor{red}{+3.9}) & 66.9 (\textcolor{red}{+10.6}) & w/ Beta-KD (Instance) & 53.8 (\textcolor{red}{+4.6}) & 65.1 (\textcolor{red}{+10.0}) \\ \hline
CE + RKL           & 46.2        & 50.7        & CE + TVD           & 49.1        & 53.1         & CE + MSE-Probs     & 47.1        & 52.1         \\
w/ Manual          & 47.8        & 53.5        & w/ Manual          & 50.1        & 56.9         & w/ Manual          & 49.3        & 56.1         \\
\rowcolor{LightCyan}
w/ Beta-KD (Task)     & 49.5 (\textcolor{red}{+1.8}) & 56.4 (\textcolor{red}{+3.0}) & w/ Beta-KD (Task)     & 51.3 (\textcolor{red}{+1.2}) & 61.0 (\textcolor{red}{+4.1})  & w/ Beta-KD (Task)     & 51.7 (\textcolor{red}{+2.4}) & 60.3 (\textcolor{red}{+4.2})  \\
\rowcolor{LightCyan}
w/ Beta-KD (Instance) & 52.4 (\textcolor{red}{+4.6}) & 61.6 (\textcolor{red}{+8.1}) & w/ Beta-KD (Instance) & 52.0 (\textcolor{red}{+1.9}) & 60.0 (\textcolor{red}{+3.1})  & w/ Beta-KD (Instance) & 52.2 (\textcolor{red}{+2.9}) & 62.4 (\textcolor{red}{+6.3})  \\ \hline
CE + SFKL          & 49.4        & 57.7        & CE + AdaptiveKL    & 49.2        & 55.2         & CE + Cosine-Logits        & 48.4        & 55.6         \\
w/ Manual          & 51.1        & 60.4        & w/ Manual          & 49.6        & 56.0         & w/ Manual          & 50.7        & 59.3         \\
\rowcolor{LightCyan}
w/ Beta-KD (Task)     & 53.1 (\textcolor{red}{+1.9}) & 63.2 (\textcolor{red}{+2.9}) & w/ Beta-KD (Task)     & 50.2 (\textcolor{red}{+0.6}) & 57.0 (\textcolor{red}{+1.0})  & w/ Beta-KD (Task)     & 53.1 (\textcolor{red}{+2.4}) & 63.3 (\textcolor{red}{+3.9})  \\
\rowcolor{LightCyan}
w/ Beta-KD (Instance) & 51.0 (-0.1) & 60.0 (-0.3) & w/ Beta-KD (Instance) & 54.2 (\textcolor{red}{+4.6}) & 65.5 (\textcolor{red}{+9.5})  & w/ Beta-KD (Instance) & 53.7 (\textcolor{red}{+3.0}) & 64.9 (\textcolor{red}{+5.6})  \\ \hline
CE + SRKL          & 48.6        & 55.2        & CE + TAID          & 46.2        & 49.3         & CE + Cosine-Probs  & 51.6        & 59.1         \\
w/ Manual          & 50.2        & 57.8        & w/ Manual          & 50.1        & 56.7         & w/ Manual          & 52.8        & 62.1         \\
\rowcolor{LightCyan}
w/ Beta-KD (Task)     & 52.0 (\textcolor{red}{+1.8}) & 60.7 (\textcolor{red}{+2.8}) & w/ Beta-KD (Task)     & 54.1 (\textcolor{red}{+4.0}) & 64.4 (\textcolor{red}{+7.7})  & w/ Beta-KD (Task)     & 54.2 (\textcolor{red}{+1.4}) & 65.3 (\textcolor{red}{+3.2})  \\
\rowcolor{LightCyan}
w/ Beta-KD (Instance) & 54.3 (\textcolor{red}{+4.1}) & 67.3 (\textcolor{red}{+9.5}) & w/ Beta-KD (Instance) & 54.8 (\textcolor{red}{+4.7}) & 65.9 (\textcolor{red}{+9.2})  & w/ Beta-KD (Instance) & \textbf{54.9} (\textcolor{red}{+2.1}) & \textbf{67.5} (\textcolor{red}{+5.5}) \\
\noalign{\hrule height 1pt} 
\end{tabular}
\end{table*}

\begin{figure}[htb!]
    \centering
    \includegraphics[width=0.47\textwidth]{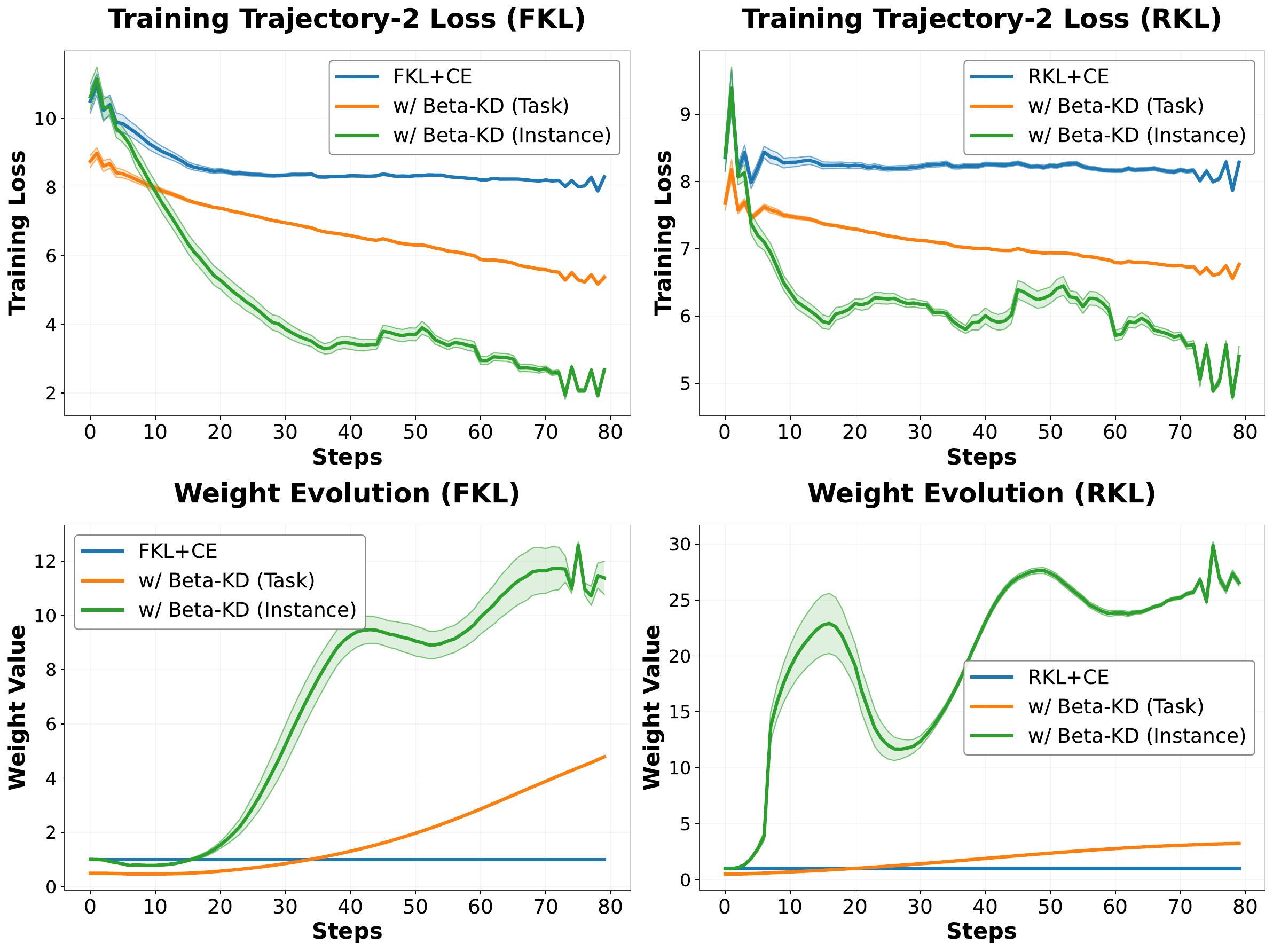}
    \caption{
    Training trajectories and dynamic weight evolution for \textbf{FKL+CE} and \textbf{RKL+CE} objectives. 
    The upper row shows the total training loss over steps, and the lower row illustrates the adaptive evolution of task and instance-level uncertainty weights $\beta$. The adaptive adjustment of the weighting parameter $\beta$ during training \textbf{ensure a faster overall loss convergence and enhances optimization stability}.
    }
    \label{fig:loss_weight}
\end{figure}

\begin{figure}[htb!]
    \centering
    \includegraphics[width=0.47\textwidth]{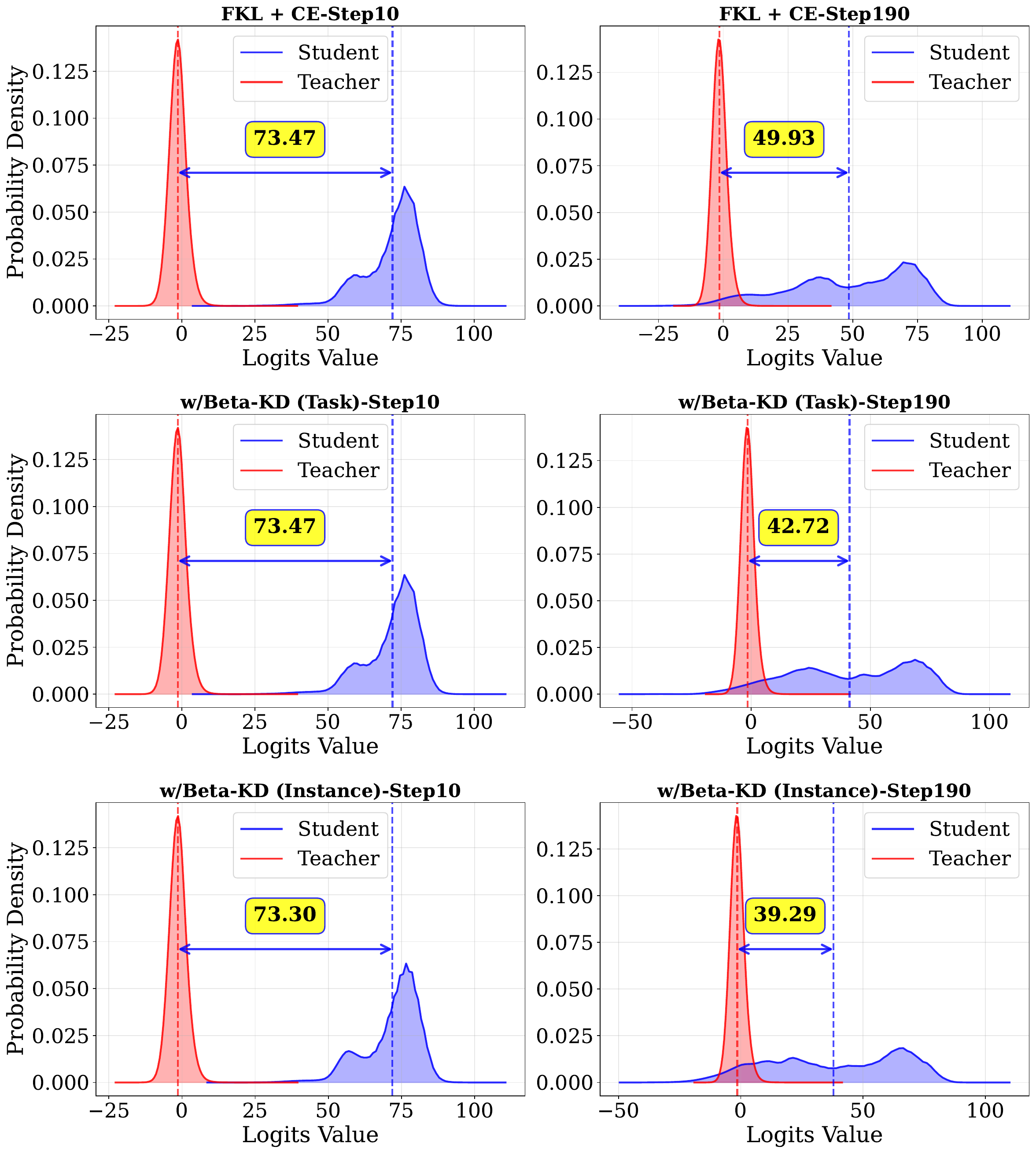}
    \caption{
    Visualization of teacher–student logit distributions at different training stages. 
    Step10 and Step190 denote early and late training checkpoints. 
    Compare with the training steps, \textbf{both Beta-KD (Task) and Beta-KD (Instance) reduce the \textbf{logit matching distance} compared to the baseline}, with the instance-level variant achieving the closest alignment.
    }
    \label{fig:logit_distribution}
\end{figure}

\begin{table*}[htbp]
\centering
\caption{
Experimental results of \textbf{three-loss balancing} on the \textbf{ScienceQA} dataset. 
Each baseline combines Cross-Entropy (CE), a KL-based distillation loss, and a feature-level distillation (FD) objective. 
\textbf{Manual} uses fixed weights among CE, KL, and FD based on their initial scales. 
\textbf{Beta-KD (Task)} models task-level uncertainty shared across all samples, 
while \textbf{Beta-KD (Instance)} models instance-level uncertainty adaptive to each input. 
}
\label{tab:scienceqa_fd_results}
\setlength{\tabcolsep}{4.5pt}
\fontsize{7}{9}\selectfont
\begin{tabular}{lll|lll|lll}
\noalign{\hrule height 1pt} 
\rowcolor{Gray}
\textbf{Method }            & \textbf{VQA-Acc}     & \textbf{IMG-Acc}     & \textbf{Method}             & \textbf{VQA-Acc}     & \textbf{IMG-Acc}      & \textbf{Method}             & \textbf{VQA-Acc}     & \textbf{IMG-Acc}      \\ \hline
CE + FKL + FD      & 45.5        & 47.9        & CE + JS + FD         & 48.4        & 53.2        & CE + MSE-Logits + FD    & 46.4        & 51.1        \\
w/ Manual          & 47.7        & 53.3        & w/ Manual            & 48.9        & 53.7        & w/ Manual               & 47.4        & 52.6        \\
\rowcolor{LightCyan}
w/ Beta-KD (Task)     & 50.2 (\textcolor{red}{+2.5}) & 59.1 (\textcolor{red}{+5.8}) & w/ Beta-KD (Task)       & 49.7 (\textcolor{red}{+0.8}) & 54.6 (\textcolor{red}{+0.9}) & w/ Beta-KD (Task)          & 48.6 (\textcolor{red}{+1.2}) & 54.5 (\textcolor{red}{+1.9}) \\
\rowcolor{LightCyan}
w/ Beta-KD (Instance) & \textbf{52.9} (\textcolor{red}{+5.2}) & \textbf{61.9} (\textcolor{red}{+8.6}) & w/ Beta-KD (Instance)   & 52.7 (\textcolor{red}{+3.8}) & 62.3 (\textcolor{red}{+8.6}) & w/ Beta-KD (Instance)      & 51.3 (\textcolor{red}{+3.9}) & 60.9 (\textcolor{red}{+8.3}) \\ \hline
CE + RKL + FD      & 50.3        & 56.5        & CE + TVD + FD        & 49.7        & 55.1        & CE + MSE-Probs + FD     & 48.2        & 54.0        \\
w/ Manual          & 49.8        & 56.0        & w/ Manual            & 51.8        & 59.7        & w/ Manual               & 48.1        & 54.5        \\
\rowcolor{LightCyan}
w/ Beta-KD (Task)     & 49.5 (-0.3) & 55.8 (-0.2) & w/ Beta-KD (Task)       & 54.1 (\textcolor{red}{+2.4}) & 64.7 (\textcolor{red}{+4.9}) & w/ Beta-KD (Task)          & 48.2 (\textcolor{red}{+0.1}) & 55.3 (\textcolor{red}{+0.8}) \\
\rowcolor{LightCyan}
w/ Beta-KD (Instance) & 50.1 (\textcolor{red}{+0.3}) & 57.5 (\textcolor{red}{+1.5}) & w/ Beta-KD (Instance)   & 49.0 (-2.8) & 56.8 (-2.9) & w/ Beta-KD (Instance)      & 47.1 (-1.0) & 53.1 (-1.4) \\ \hline
CE + SFKL + FD     & 51.5        & 59.0        & CE + AdaptiveKL + FD & 47.4        & 51.5        & CE + Cosine-Logits + FD & 44.9        & 48.4        \\
w/ Manual          & 51.3        & 59.1        & w/ Manual            & 49.7        & 56.6        & w/ Manual               & 48.1        & 55.0        \\
\rowcolor{LightCyan}
w/ Beta-KD (Task)     & 51.4 (\textcolor{red}{+0.1}) & 59.6 (\textcolor{red}{+0.5}) & w/ Beta-KD (Task)       & 52.3 (\textcolor{red}{+2.6}) & 62.1 (\textcolor{red}{+5.4}) & w/ Beta-KD (Task)          & 51.5 (\textcolor{red}{+3.4}) & 62.0 (\textcolor{red}{+6.9}) \\
\rowcolor{LightCyan}
w/ Beta-KD (Instance) & 50.0 (-1.3) & 59.1 (-0.1) & w/ Beta-KD (Instance)   & 49.5 (-0.2) & 56.4 (-0.2) & w/ Beta-KD (Instance)      & 51.4 (\textcolor{red}{+3.3}) & 60.1 (\textcolor{red}{+5.1}) \\ \hline
CE + SRKL + FD     & 48.3        & 52.4        & CE + TAID + FD       & 48.3        & 54.4        & CE + Cosine + FD        & 47.7        & 52.3        \\
w/ Manual          & 49.4        & 55.1        & w/ Manual            & 49.0        & 55.9        & w/ Manual               & 49.4        & 56.9        \\
\rowcolor{LightCyan}
w/ Beta-KD (Task)     & 50.9 (\textcolor{red}{+1.5}) & 58.2 (\textcolor{red}{+3.1}) & w/ Beta-KD (Task)       & 50.0 (\textcolor{red}{+1.0}) & 57.7 (\textcolor{red}{+1.8}) & w/ Beta-KD (Task)          & 51.3 (\textcolor{red}{+1.9}) & 61.8 (\textcolor{red}{+4.9}) \\
\rowcolor{LightCyan}
w/ Beta-KD (Instance) & 53.9 (\textcolor{red}{+4.5}) & 63.3 (\textcolor{red}{+8.1}) & w/ Beta-KD (Instance)   & 53.2 (\textcolor{red}{+4.2}) & 64.8 (\textcolor{red}{+8.9}) & w/ Beta-KD (Instance)      & \textbf{54.2} (\textcolor{red}{+4.8}) & \textbf{64.8} (\textcolor{red}{+7.9}) \\
\noalign{\hrule height 1pt} 
\end{tabular}
\end{table*}

\subsection{Effectiveness of Uncertainty Weighting}

\begin{table*}[htbp]
\centering
\caption{
\textbf{Experimental results of the proposed uncertainty weighting framework on multiple benchmarks.} 
$\mathrm{MME}^A$ is obtained by dividing $\mathrm{MME}^P$ by 20 to align its scale with other benchmarks. 
The overall average (Avg.) is computed across six datasets: $\mathrm{MME}^P$, GQA, $\mathrm{VQA}^T$, POPE, $\mathrm{MMB^{dev}}$, and $\mathrm{SQA}^I$. 
\textbf{Align-KD}'s result is our reproduced result, while \textbf{Cosine KD} means replaces the FKL loss in Align-KD with our proposed \textit{probability-space Cosine Distance Matching}. 
Our uncertainty-aware Beta-KD framework further improves both variants, achieving new state-of-the-art performance.
}

\label{tab:6-task}
\setlength{\tabcolsep}{2pt}
\fontsize{7}{9}\selectfont
\begin{tabular}{lll|lllllllll}
\noalign{\hrule height 1pt}
\rowcolor{Gray}
\textbf{Method}\rule{0pt}{1.4em} & \textbf{LLM} & \textbf{\#Samples} &
\textbf{MME$^{P}$} & \textbf{MME$^{A}$} & \textbf{GQA} &
\textbf{VQA$^{T}$} & \textbf{POPE} & \textbf{MMB$^{dev}$} &
\textbf{SQA$^{I}$} & \textbf{Avg.} \\ \hline
LLaVA-1.5 \cite{llava-1.5}         & Vicuna-7B        & 1.2M      & 1510.7 & 75.5 & 62.0 & 58.2 & 85.9 & 64.3 & 66.8 & 68.8 \\
ShareGPT4V \cite{sharegpt4v}        & Vicuna-7B        & 1.9M      & 1567.4 & 78.4 & 63.3 & 60.4 & 85.7 & 68.8 & 68.4 & 70.8 \\
MobileVLM V2 7B \cite{mobilevlm-v2}   & Vicuna-7B        & 3.6M      & 1559.0 & 78.0 & 62.6 & 62.3 & 86.6 & 69.2 & 74.7 & 72.2 \\ \hline
MoE-LLaVA-2B \cite{mobilevlm}      & Qwen-1.5-1.8B    & 2.2M      & 1292.0 & 64.6 & 61.5 & 48.0 & 87.0 & 59.7 & 63.1 & 64.0 \\
MoE-LLaVA-1.6B \cite{moe-llava}    & StableLM-1.6B    & 2.2M      & 1300.0 & 65.0 & 60.4 & 47.8 & 84.3 & 59.4 & 62.6 & 63.3 \\ 
MobileVLM 3B \cite{mobilevlm}      & MobileLLaMA 2.7B & 3.6M      & 1288.9 & 64.4 & 59.0 & 47.5 & 84.9 & 59.6 & 61.2 & 62.8 \\ \hline
\rowcolor{LightGray}
MobileVLM 1.7B \cite{mobilevlm}    & MobileLLaMA 1.4B & 3.6M      & 1196.2 & 59.8 & 56.1 & 41.5 & 84.5 & 53.2 & 57.3 & 58.7 \\
\rowcolor{LightGray}
MobileVLM -V2 1.7B \cite{mobilevlm-v2} & MobileLLaMA 1.4B & 3.6M      & 1246.3 & 62.3 & 55.1 & 51.2 & 85.3 & 57.6 & 63.2 & 62.5 \\
\rowcolor{LightGray}
LLAVADI \cite{llavadi}           & MobileLLaMA 1.4B & 3.6M      & 1178.6 & 58.9 & 55.4 & 45.3 & 84.7 & 55.0 & 56.0 & 59.2 \\ \hline
\rowcolor{LightGray}
Align-KD \cite{alignkd}          & MobileLLaMA 1.4B & 3.6M      & 1288.9 & 64.4 & 60.0 & 52.8 & 85.4 & 57.0 & 61.0 & 63.4 \\
\rowcolor{LightCyan}
w/ Beta-KD (Task)     & MobileLLaMA 1.4B & 3.6M      & 1315.1 (\textcolor{red}{+26.3}) & 65.8 (\textcolor{red}{+1.3}) & 60.2 (\textcolor{red}{+0.2}) & 52.9 (\textcolor{red}{+0.1}) & 85.4 (+0.0) & 58.8 (\textcolor{red}{+1.8}) & 61.4 (\textcolor{red}{+0.4}) & 64.1 (\textcolor{red}{+0.6}) \\
\rowcolor{LightCyan}
w/ Beta-KD (Instance) & MobileLLaMA 1.4B & 3.6M      & 1343.0 (\textcolor{red}{+54.1}) & 67.1 (\textcolor{red}{+2.7}) & 60.8 (\textcolor{red}{+0.8}) & 53.9 (\textcolor{red}{+1.1}) & 85.4 (+0.0) & 59.1 (\textcolor{red}{+2.2}) & 61.2 (\textcolor{red}{+0.2}) & 64.6 (\textcolor{red}{+1.2}) \\ \hline
\rowcolor{LightGray}
Cosine KD          & MobileLLaMA 1.4B & 3.6M      & 1308.4 & 65.4 & 59.9 & 52.2 & 84.6 & 57.1 & 61.3 & 63.4 \\
\rowcolor{LightCyan}
w/ Beta-KD (Task)     & MobileLLaMA 1.4B & 3.6M      & 1352.0 (\textcolor{red}{+43.6}) & 67.6 (\textcolor{red}{+2.2}) & 60.8 (\textcolor{red}{+0.9}) & 53.9 (\textcolor{red}{+1.7}) & 85.4 (\textcolor{red}{+0.8}) & 59.1 (\textcolor{red}{+2.0}) & 61.2 (-0.1) & 64.7 (\textcolor{red}{+1.2}) \\
\rowcolor{LightCyan}
w/ Beta-KD (Instance) & MobileLLaMA 1.4B & 3.6M      & \textbf{1350.3} (\textcolor{red}{+42.0}) & \textbf{67.5} (\textcolor{red}{+2.1}) & \textbf{61.2} (\textcolor{red}{+1.3}) & \textbf{54.2} (\textcolor{red}{+1.9}) & \textbf{86.6} (\textcolor{red}{+1.9}) & \textbf{60.2} (\textcolor{red}{+3.1}) & \textbf{63.1} (\textcolor{red}{+1.8}) & \textbf{65.5} (\textcolor{red}{+2.0}) \\
\noalign{\hrule height 1pt}
\end{tabular}
\end{table*}

\textbf{Results.}  
The experimental results of combining CE and various KL-based distillation losses are summarized in Table~\ref{tab:scienceqa_uncertainty} and Table~\ref{tab:scienceqa_fd_results}. 
From the results, we draw three key conclusions. 
\textbf{(1) The proposed uncertainty-aware adaptive weighting (\textbf{Beta-KD}) consistently outperforms both manual weighting and the original unweighted baseline (CE + KL) or (CE + KL + FD).} 
Across all distillation objectives, both task-level and instance-level Beta-KD significantly improve VQA and IMG accuracy by approximately +1$\sim$5\% compared to manual tuning, demonstrating that Bayesian uncertainty-based weighting effectively balances the contributions of multiple objectives. 
\textbf{(2) Instance-level Beta-KD consistently outperforms task-level Beta-KD, with an average improvement of +2$\sim$6\%.} 
This indicates that dynamically adjusting weights at the sample level better captures data heterogeneity and noise, leading to more stable optimization and stronger generalization. 
\textbf{(3) Among all variants, \textbf{Cosine-Probs} achieve the best results under Beta-KD, showing strong robustness in the probability space.} 
In particular, CE + Cosine-Probs (Instance) achieves the highest VQA-Acc (54.9\%) and IMG-Acc (67.5\%), surpassing all KL variants. 
This suggests that direction-based matching in the post-softmax probability space, when combined with uncertainty weighting, better preserves the relative probability structure between teacher and student models, leading to superior multimodal knowledge distillation performance.

Compared to Kendall \emph{et al.}'s work \cite{multitask}, who model uncertainty at the \emph{task level}, we explicitly capture \emph{instance-level uncertainty}. As shown in Figure~\ref{fig:logit_bar}, during early training, high student entropy indicates greater uncertainty, and the distillation loss is assigned a larger weight (e.g., $1.98$), encouraging stronger guidance from the teacher. As training progresses, student entropy decreases and confidence increases, leading to a reduced distillation weight (e.g., $0.3$) and greater reliance on the student’s own predictions. These results demonstrate that our \emph{instance-level weighting strategy} adaptively allocates importance at the sample level.

\begin{figure}[htb!]
    \vspace{-6pt}
    \centering
    \setlength{\abovecaptionskip}{2pt}
    \setlength{\belowcaptionskip}{0pt}
    \includegraphics[width=0.4\textwidth]{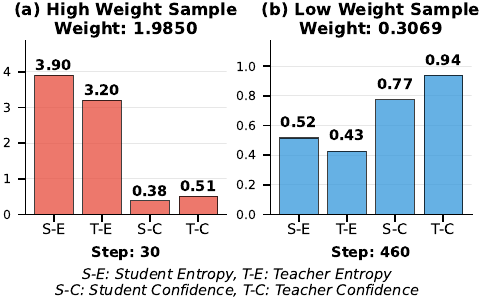}
    \caption{Visualization of Student Entropy}
    \label{fig:logit_bar}
    \vspace{-6pt}
\end{figure}

\subsection{How Uncertainty Weighting Works}



\textbf{Figure~\ref{fig:loss_weight} shows that uncertainty-based weighting yields faster and more stable optimization.}
Both Beta-KD (Task) and Beta-KD (Instance) converge faster and achieve lower final losses than the fixed-weight baseline.
Meanwhile, the learned weights adapt to uncertainty over training, with instance-level Beta-KD showing stronger dynamics.

\textbf{Figure~\ref{fig:logit_distribution} shows that uncertainty weighting improves student--teacher alignment.}
We compare student and teacher logit distributions at early and late stages and measure the \emph{matching distance} as the mean difference between the two distributions.
Beta-KD (Task/Instance) consistently reduces this distance, indicating improved student--teacher consistency.

\subsection{Generalization Ability}

\begin{table}[htbp]
\vspace{-5pt}
\centering
\caption{
Performance comparison on LLAVA-Qwen Structure on \textbf{TextVQA} and \textbf{ScienceQA}.
We report VQA accuracy (\%).}
\label{tab:aukd_results}
\setlength{\tabcolsep}{5pt}
\fontsize{7}{9}\selectfont
\begin{tabular}{ll|cc|cc}
\noalign{\hrule height 1pt}
\textbf{Model Size} & \textbf{Method} 
& \multicolumn{2}{c|}{\textbf{TextVQA}} 
& \multicolumn{2}{c}{\textbf{ScienceQA}} \\ 
& & Acc. & Gain & Acc. & Gain \\ \hline

\multicolumn{6}{l}{\textbf{0.5B Model (Qwen2.5-0.5B)}} \\ \hline
\rowcolor{LightGray}
LLaVA-KD               & KD             
& 52.0 & -- 
& 60.6 & -- \\
\rowcolor{LightCyan}
w/ Beta-KD(Task)         & Task-level     
& 53.1 & (\textcolor{red}{+1.1}) 
& 61.9 & (\textcolor{red}{+1.3}) \\
\rowcolor{LightCyan}
w/ Beta-KD(Instance)     & Instance-level 
& \textbf{54.9} & (\textcolor{red}{+2.9}) 
& \textbf{64.4} & (\textcolor{red}{+3.8}) \\ \hline

\multicolumn{6}{l}{\textbf{1.5B Model (Qwen2.5-1.5B)}} \\ \hline
\rowcolor{LightGray}
LLaVA-KD               & KD             
& 59.7 & -- 
& 71.4 & -- \\
\rowcolor{LightCyan}
w/ Beta-KD (Task)         & Task-level      
& 61.0 & (\textcolor{red}{+1.3}) 
& 71.9 & (\textcolor{red}{+0.5}) \\
\rowcolor{LightCyan}
w/ Beta-KD (Instance)     & Instance-level 
& \textbf{62.7} & (\textcolor{red}{+3.0}) 
& \textbf{73.5} & (\textcolor{red}{+1.9}) \\

\noalign{\hrule height 1pt}
\end{tabular}
\label{table:qwen}
\end{table}
We re-implement both our task-level and instance-level weighting strategies within the LLaVA-style architecture. Using Qwen2.5-3B as a teacher model, We observe that our method consistently improves distillation performance on both 1.5B and 0.5B student models. These results further confirm the effectiveness of our approach beyond the MobileVLM architecture.

\subsection{Computational complexity analysis}
Task-level uncertainty introduces only three scalar log-variance terms, while instance-level uncertainty employs a lightweight two-layer MLP with just \textbf{0.03\%} parameters of the 1.67B-parameter backbone. The additional memory overhead is negligible. As shown in Table~\ref{tab:training_speed}, all weighting strategies exhibit nearly identical training speed and memory usage, since the weighting computation incurs less than 0.01\% of total training time, which is dominated by student and teacher forward passes.


\begin{table}[htbp]
\vspace{-8pt}
\centering
\setlength{\abovecaptionskip}{2pt}
\setlength{\belowcaptionskip}{0pt}
\setlength{\tabcolsep}{0.5pt}
\fontsize{7}{9}\selectfont
\caption{Statistics of training efficiency.}
\label{tab:training_speed}
\begin{tabular}{lccccc}
\toprule
\textbf{Strategy}
& \textbf{Speed (it/s)}
& \textbf{Time/Epo.}
& \textbf{GPU Mem.}
& \textbf{Params}
& \textbf{Mem.} \\
\midrule
\rowcolor{LightGray}
Align-KD                     & 1.82 & 11:35 min & 47.5 GB & 0     & 0 MB \\
\rowcolor{LightCyan}
w/ Beta-KD (Task)         & 1.87 & 11:30 min & 47.5 GB & 3     & $<$1 MB \\
\rowcolor{LightCyan}
w/ Beta-KD (Instance)     & 1.85 & 11:35 min & 47.6 GB & 524K  & $\sim$4 MB \\
\bottomrule
\end{tabular}
\vspace{-8pt}
\end{table}

\section{Conclusion}
{We present} Beta-KD, a unified Bayesian framework for uncertainty-aware multimodal distillation.  
By reformulating multi-channel knowledge transfer as a Bayesian inference {problem} under a teacher-informed Gibbs prior {on student activations}, Beta-KD interprets each supervision channel through a precision parameter $\beta$, which quantifies the reliability of teacher guidance.  
Extensive experiments demonstrate that Beta-KD not only stabilizes training and improves generalization across diverse multimodal benchmarks, but also achieves consistent gains over strong baselines, providing a scalable and theoretically grounded approach for learning from imperfect multimodal data and model supervisions.

\paragraph{Acknowledgement} This work is partially supported by NSF AI Institute2229873, NSF RI-2223292, an Amazon research award, and
an Adobe gift fund.
 \small \bibliographystyle{ieeenat_fullname} \bibliography{main}
}


\end{document}